\documentclass[12pt]{article}
\usepackage{amsmath}
\usepackage{bm}
\usepackage{graphicx}
\usepackage{color}
\usepackage[round]{natbib}
\usepackage{textcomp}
\usepackage{subfigure}   

\begin{document}

\def\btheta{{\boldsymbol{\theta}}}
\def\bfphi{{\boldsymbol{\phi}}}
\def\bfvarphi{{\boldsymbol{\varphi}}}
\def\bmu{{\boldsymbol{\mu}}}
\def\bY{{\mathbf Y}}
\def\bA{{\mathbf A}}
\def\bB{{\mathbf B}}
\def\bY{{\mathbf Y}}
\def\bN{{\mathbf N}}
\def\bH{{\mathbf H}}
\def\bq{{\mathbf q}}
\def\bQ{{\mathbf Q}}
\def\bP{{\mathbf P}}
\def\bx{{\mathbf x}}
\def\bfY{{\mathbf Y}}
\def\argmax{{\mathrm{argmax}}}

\renewcommand{\thefootnote}{$\text{c1}$}

\begin{center}\textbf{\large A Dictionary Approach to EBSD Indexing}\\
\vspace*{0.25in}[brief title: \textit{Dictionary Approach to EBSD Indexing}]
\end{center}
\vspace*{0.25in}
\begin{center}
Yu-Hui Chen$^{a1}$, Se Un Park$^{a2}$, Dennis Wei$^{a3}$,  Gregory Newstadt$^{a1}$, Michael Jackson$^{a4}$, Jeff P. Simmons$^{a5}$, Marc De Graef$^{a6}$\footnote{Corresponding Author.  E-mail: degraef@cmu.edu}, and Alfred O. Hero$^{a1}$
\end{center}
\vspace*{0.25in}

\noindent $^{\text{a1}}$ Department of Electrical Engineering and Computer Science, University of Michigan, Ann Arbor, MI 48109, USA

\noindent $^{\text{a2}}$ Schlumberger Research, Cambridge, MA USA

\noindent $^{\text{a3}}$ IBM Watson Research Center, Yorktown Heights, NY, USA

\noindent $^{\text{a4}}$ Bluequartz Software, Dayton, OH, USA

\noindent $^{\text{a5}}$ Materials and Manufacturing Directorate, AFRL/MLLMD, Wright-Patterson AFB, OH 45433, USA

\noindent $^{\text{a6}}$ Department of Materials Science and Engineering, Carnegie Mellon University, 5000 Forbes Avenue, Pittsburgh PA 15213, USA

\newpage
\section*{Abstract}

We propose a framework for indexing of grain and sub-grain structures in electron backscatter diffraction (EBSD) images of polycrystalline materials. The framework is based on a previously introduced physics-based forward model by~\citet{callahan_dynamical_2013} relating measured patterns to grain orientations (Euler angle). The forward model is tuned to the microscope and the sample symmetry group. We discretize the domain of the forward model onto a dense grid of Euler angles and for each measured pattern we identify the most similar patterns in the dictionary. These patterns are used to identify boundaries, detect anomalies, and index crystal orientations. The statistical distribution of these closest matches is used in an unsupervised binary decision tree (DT) classifier to identify grain boundaries and anomalous regions. The DT classifies a pattern as an anomaly if it has an abnormally low similarity to any pattern in the dictionary. It classifies a pixel as being near a grain boundary if the highly ranked patterns in the dictionary differ significantly over the pixel’s $3\times3$ neighborhood. Indexing is accomplished by computing the mean orientation of the closest dictionary matches to each pattern. The mean orientation is estimated using a maximum likelihood approach that models the orientation distribution as a mixture of Von Mises–Fisher distributions over the quaternionic $3$-sphere. The proposed dictionary matching approach permits segmentation, anomaly detection, and indexing to be performed in a unified manner with the additional benefit of uncertainty quantification. We demonstrate the proposed dictionary-based approach on a Ni-base IN100 alloy. \footnote{
Part of this work was reported in the Proceedings of the IEEE International Conference on Image Processing (ICIP), Melbourne Australia, Sept 2013.}
\vspace*{0.25in}
\section*{Keywords}
electron back-scatter diffraction pattern;\\ EBSD;\\ dynamical electron scattering;\\ dictionary matching;\\ maximum likelihood orientation estimates;\\
Von Mises--Fisher mixture distribution

\section{Introduction}
\label{sec:intro}
Electron backscatter diffraction, EBSD, is used to perform quantitative microstructure analysis of polycrystalline materials on a millimeter to nanometer scale (\citet{schwartz_electron_2009}). Most current EBSD segmentation (i.e., delineation of individual grains by determination of the grain boundary locations) and indexing (i.e., orientation determination) methods extract orientations and widths of Kikuchi bands in a measured pattern by using a modified Hough transform, implemented using image processing tools such as butterfly convolution, Gaussian filtering, binning, peak detection, and image quality maps to gauge indexing and segmentation accuracy (\citet{tao_errors_2005}), (\citet{wright_ebsd_2006}). By comparing the measured diffraction line parameters to a pre-computed database, indexing yields the crystal orientation, commonly described by three Euler angles with respect to a reference frame, for the volume illuminated by the beam. By repeating the process on a grid of scanning locations on the sample, an orientation map or image is produced. The image is then segmented into grains by thresholding normed differences between the Euler angles (misorientations). The accuracy of the Hough approach to EBSD indexing depends to a large extent on the visibility of the Kikuchi bands, which is often represented in terms of an "image quality" parameter.

In this paper we propose an alternative indexing approach that uses a physics-based forward model for the full diffraction patterns and does not require application of the Hough transform or other image processing tools. The proposed approach exploits the known physics of electron scattering phenomena that underlies the image formation process. With this additional information, grain boundaries and anomalous points can be detected as explicit classes at the same time as grains are segmented. Anomaly detection (i.e., the automated detection of abnormal or unexpected diffraction patterns) is an important capability, since anomalies may correspond to defects or contaminants that affect the material properties. In addition, unlike methods based on the Hough transform, the proposed pattern dictionary approach differentiates grain interiors from grain boundaries without requiring additional processing of the measured patterns.

Automated indexing of electron diffraction patterns by means of pattern matching techniques is not new, and was proposed in 1991 by Wright and coworkers (\citet{wright_automated_1991}) for backscattered Kikuchi diffraction patterns, now commonly known as electron backscatter diffraction patterns or EBSPs. Their approach involved automated comparisons between a series of experimental patterns and idealized patterns, created from a set of orientations that uniformly covers the asymmetric region (or fundamental zone) of Euler space. While the technique showed promising results, computer limitations prevented the approach from gaining widespread acceptance. In the context of precession electron diffraction in the transmission electron microscope, Rauch and coworkers (\citet{rauch_rapid_2005,rauch_automatic_2008}) proposed a template-based pattern matching approach for the automatic orientation determination of quasi-kinematical diffraction patterns. The templates are computed from kinematical structure factors, and the proper asymmetric part of Euler space is sampled uniformly to generate a template collection, which is then compared quantitatively against the experimental patterns. In this template matching process, only the top match is considered in the determination of the crystal orientation. As we will explain in detail in the present paper, our dictionary appproach uses, in addition to a physics-based model for the generation of the dictionary, a statistical model for the orientation distribution of all the highly ranked pattern matches to provide both a stable and robust estimate of orientation, as well as a quantitative statistical uncertainty; the method proposed in~\citet{rauch_rapid_2005} does not perform such a statistical analysis. The current commercially available EBSD indexing suites also lack a statistical determination of the uncertainties in the orientation determination.

The proposed indexing framework relies on two components: offline dictionary generation and online dictionary matching to the sample. Offline dictionary generation is accomplished as follows. First, a dictionary of raw diffraction patterns is generated using the forward model of~\citet{callahan_dynamical_2013}. This dictionary is tuned to the parameters of the microscope and the crystal symmetry group(s) of the sample. Second, the singular value decomposition (SVD) of the dictionary is computed and a second dictionary, called the dictionary of background compensated patterns, is generated by projection of the raw dictionary onto the space orthogonal to the first principal component; this is essentially a background subtraction process.

The first step in the online dictionary matching algorithm is to compute normalized inner products between the uncompensated (raw) sample patterns and the uncompensated dictionary. This is repeated on the compensated sample patterns and the compensated dictionary patterns. The basis for the online dictionary matching algorithm is the construction of a pair of bipartite graphs, uncompensated and compensated, respectively, using these normalized inner products. A bipartite graph is a graph connecting vertices or nodes in two disjoint sets (\citet{diestel_graph_2005}); in our case, the first set contains the experimental patterns at different locations on the sample, the second the dictionary patterns for different orientations. For each spatial location on the sample, the top $k$ (normalized) inner products between the sample diffraction pattern and a dictionary determine the $k$-nearest-neighbor (kNN) neighborhood of the pattern in the dictionary. These kNN neighborhoods form the backbone of the proposed online dictionary matching algorithm.

From the bipartite graphs, sample patterns can be classified as grain interiors, grain
boundaries, or anomalies, using an unsupervised decision tree (DT) classifier defined on the graphs. Specifically, the classifier uses the shapes of these kNN neighborhoods to discriminate between these types of patterns. A tightly clustered and connected kNN neighborhood indicates a grain interior sample pattern. A kNN neighborhood that forms two or three clusters in the dictionary indicates a grain boundary sample pattern. kNN neighborhoods that are very spread-out and scattered indicate anomalous sample patterns. The unsupervised DT classifier uses the uncompensated dictionary to distinguish unusual (anomalous) background patterns. The compensated dictionary is used to distinguish non-anomalous patterns as interior to a grain or on the boundary of a grain. A pixel is classified as "anomalous", "grain interior" or "grain boundary" using an unsupervised decision tree (DT) classifier to test homogeneity of the pattern matches over a $3\times3$ spatial patch centered at the pixel. The effect of surface roughness, and the resulting shifts in the EBSD background intensity, is a topic of ongoing analysis.

Indexing of crystal orientation is performed using a maximum likelihood (ML) estimation strategy for determining the Euler angles. The ML strategy fits a Von Mises-Fisher mixture (VMFm) density model to the observed distribution of the Euler angles of the top dictionary matches; the von Mises-Fisher distribution is a probability distribution on a sphere in a $p$-dimensional space. An iterative expectation-maximization (EM) estimation algorithm is used to perform the fit. The use of the VMFm model allows one to account for the symmetry-induced ambiguities of the crystal orientation and produces an estimate of the Euler angles in the desired fundamental zone. A side benefit is that the algorithm yields an estimate of the spread of the VMFm model that can be used as an a posteriori confidence measure on the orientation estimate.

To the best of our knowledge, the framework proposed here is the first EBSD indexing approach that uses a dictionary generated by a physics-based forward model. Some advantages of our model-based approach are: 1) it incorporates the physics of dynamical electron scattering; 2) it unifies segmentation, indexing and anomaly detection; 3) it incorporates a statistical model that naturally generates both an estimate of orientation and a measure of confidence in the estimate; 4) it involves parallelizable operations relying on simple inner products and nearest neighbor search. At the same time, the large size of the dictionary, together with the high dimension of the diffraction patterns, create computational challenges as discussed in Sec.~\ref{sec:computation}.

The outline of the chapter is as follows. Section~\ref{sec:Dict_model} presents the proposed dictionary model for EBSD pattern classification and indexing. Section~\ref{sec:classification_indexing} develops the statistical algorithms for anomaly detection, segmentation, and indexing based on the dictionary model. Section~\ref{sec:Dict_generation} describes the dictionary generation process. Section~\ref{sec:exp_methods} presents the experimental methods for generating the EBSD samples used in Sec.~\ref{sec:results}. Section~\ref{sec:results} presents the results of applying the proposed dictionary-based classification and indexing to a Ni-base IN100 alloy. Finally, Section~\ref{sec:conclusion} summarizes the paper and points to future directions.

\section{Dictionary Model}
\label{sec:Dict_model}
This section describes the non-linear forward model of~\citet{callahan_dynamical_2013}. It then shows how a physics-based dictionary of diffraction patterns can be used to approximately linearize the forward model. For descriptive economy, throughout this paper we denote by a "pixel" a particular scan location on the sample surface; with each pixel, there is an associated EBSD pattern acquired at the sample location.

\subsection{Forward model}
\label{subsec:forward_model}
When the beam is focused on a grain within the sample, the measured backscatter diffraction pattern, $\bY$, on the detector surface can be expressed as a function of the crystal orientation $\btheta$, parameterized by an Euler angle triplet $(\varphi_1, \Phi, \varphi_2)$, the incident electron energy, $E$, and interaction depth, $z_0$, as:
\begin{equation}
\label{eq:EBSD_pattern_forward_model}
\bY = \bH(\mathcal{P}(\btheta,E,z_0))+\bN,
\end{equation}
where $\bH(\mathcal{P}(\btheta,E,z_0))$ is a forward model for the backscatter process and $\bN$ is detector noise. Here $\bH$ is a measurement operator that accounts for the instrument geometry and sensitivity, and $\mathcal{P}$ represents the thickness and energy averaged mean backscatter yield. An accurate forward model for this yield was proposed by~\citet{callahan_dynamical_2013}. The model employs Monte Carlo simulations to obtain the energy, spatial, and exit depth distributions of the backscattered electrons. This statistical information is then used to compute a series of dynamical "EBSD master patterns" as a function of the electron exit energies and directions. The master pattern represents all possible EBSD patterns for a given exit energy, and specification of the grain orientation $\btheta$ along with the detector geometry then leads to an actual EBSD pattern for that orientation. These three steps constitute the forward model $\mathcal{P}$. The measurement operator $\bH$ includes the scintillator-to-CCD conversion process in the form of a point spread function for the coupling optics, as well as detector quantum efficiency and CCD binning mode. In the remainder of this paper, we will refer to the depth and energy averaged diffraction pattern generated by the forward model described above as the mean diffraction pattern.

As the electron beam is scanned across the sample surface, the diffraction pattern will change due to changes in the local crystal orientation $\btheta$ between homogeneous grains. In grain interiors, the forward model (\ref{eq:EBSD_pattern_forward_model}) can be used to produce estimates of crystal orientation at each scan location. Elsewhere in the sample, e.g., near boundary regions or near locations of anomalous features, the model (\ref{eq:EBSD_pattern_forward_model}) will no longer be a good fit to the measured patterns. Thus, the goodness of fit of the forward model can be used to classify grains, grain boundaries, and anomalies in the sample. This forms the basis for our proposed use of the forward-model to perform classification and indexing.

\subsection{Sparse dictionary-based forward model}
In principle one could formulate classification and indexing as a non-linear inverse problem using the full forward model (\ref{eq:EBSD_pattern_forward_model}). For example, given a noise model for $\bN$ one could perform maximum likelihood estimation to solve the indexing problem and likelihood-ratio testing to solve the classification problem. In the special case of a Gaussian noise model both solutions would require solving the non-linear least-squares problem $\min_\btheta\|\bY-\bH(\mathcal{P}(\btheta,E,z_0))\|^2$, in which the Euclidean norm squared of the residual fitting errors is to be minimized with respect to the orientation $\btheta$. Here, we take a simpler approach to the inverse problem that leads directly to tractable indexing and classification algorithms.

Let $\mathcal{D}$ denote a precomputed dictionary of mean diffraction patterns obtained by densely sampling the function $\bH(\mathcal{P}(\btheta,E,z_0))$ over the range of orientations $\btheta$, keeping $E$ and $z_0$ fixed. Assume that the size of $\mathcal{D}$ is $d$. Then, for sufficiently dense samples, the model (\ref{eq:EBSD_pattern_forward_model}) can be approximated by
\begin{equation}
\label{eq:EBSD_forward_model_approx}
\bY=\sum_{i=1}^dx_i\phi_i+\bN,
\end{equation}
where $\phi_i\in\mathcal{D}$ are dictionary elements and $x_i$ are coefficients. When $\bY$ corresponds to the measured pattern at a location within a grain, one might expect that only a few $x_i$'s will be non-zero, i.e., the representation (\ref{eq:EBSD_forward_model_approx}) is sparse. In particular, as the dictionary becomes increasingly dense the sparsity of the representation will also increase and, in the limit, $x_i$ will become a delta function $x_i=\Delta(i-j)$\footnote{The Delta function $\Delta(i)$ is equal to $1$ if $i=0$ and is equal to $0$ for other $i$.}, where $j$ is the index of the true orientation at that location. Note that when $\bN=0$, in this limiting case (of a fully dense dictionary) errorless estimation of $\btheta$ can be accomplished by finding the index $i$ which yields the largest normalized inner product $\rho(\bY,\phi_i)=\frac{(\bY,\phi_i)}{\|\bY\|\|\phi_i\|}$, where, for two diffraction pattern $\bA=((A_{lm}))$ and $\bB=((B_{lm}))$, $\langle\bA,\bB\rangle=\sum_{l,m}A_{lm}B_{lm}$, where $l$ and $m$ index the vertical and horizontal locations on the photodetector. The significance of this fact is that we have simplified the solution of a complicated non-linear least squares problem to the solution of a linear least squares problem followed by a table lookup (matching an index of the dictionary to the associated Euler angle).
 
In the practical case of a finite dictionary there will not be an exact match to the true diffraction pattern and (\ref{eq:EBSD_forward_model_approx}) is interpreted as a model that interpolates over the patterns in the dictionary, with interpolation coefficients $\{x_i\}_{i=1}^d$. For the purposes of indexing and classification we will restrict ourselves to sparse models, where only a few ($k$) of these coefficients are non-zero, i.e., using only a small number of dictionary elements to fit (\ref{eq:EBSD_forward_model_approx}). This sparse approximation problem is a well studied mathematical problem with many different iterative algorithms available for identifying the few non-zero coefficients. The brute force algorithm that tries to find the best fit over any set of $k$ dictionary elements is intractable except for very small values of $k$. Alternatives include basis pursuit methods such as orthogonal matching pursuit (OMP), stepwise OMP (stOMP), compressive sampling OMP, iterative soft thresholding (IST), and convex optimization relaxation methods such as $l_1$ minimization using active set, interior point, or subgradient methods (\citet{tropp_computational_2010}). In the EBSD application, the sparse approximation has to be performed for each and every pattern measured on the sample. Even with a relatively modest dictionary size and low sample resolution, e.g., $d = 100,000$ elements and $n = 100,000$ scan locations, these methods are computationally heavy.

We have adopted a simpler correlation matching approach with significantly reduced computation requirements. Instead of fitting the model (\ref{eq:EBSD_forward_model_approx}) through least squares we simply use inner products to find the $k$ top matches between the dictionary and the observations. Specifically, for each measured pattern $\bY$ we compute the normalized inner products (correlation) $\rho(\bY,\phi_1),...,\rho(\bY,\phi_d)$ between $\bY$ and the dictionary and rank them in order of decreasing magnitude. The top correlation matches are the $k$ dictionary patterns having the highest inner products\footnote{Note that these top $k$ inner product matches will be identical to the dictionary elements selected by $k$ iterations of OMP in the case that the dictionary is orthogonal.}. These $k$ patterns constitute and pixel's $k$-nearest-neighbor ($k$-NN) neighborhood in the dictionary. The collections of $k$-NN's define a bipartite graph connecting measured patterns to patterns in the dictionary. In general, these connections will be one-to-many, i.e., for each experimental pattern there will be a small number of near matches in the dictionary. The $k$-NN neighborhoods will be used to perform classification and indexing as described in Sec.~\ref{sec:classification_indexing}.

In addition to the dictionary $\mathcal{D}$, referred to as the uncompensated dictionary, we will use a derived dictionary of compensated patterns $\mathcal{D}_c$ to cluster the observed patterns into classes that can then be used for anomaly detection, segmentation and indexing. The compensated dictionary $\mathcal{D}_c$ consists of patterns in $\mathcal{D}$ after projecting away the background. This is performed by applying the singular value decomposition to determine the principal component, which closely resembles the population mean of the patterns in $\mathcal{D}$, and projecting the dictionary onto the space orthogonal to the principal component (i.e., removing the mean pattern from each individual pattern). This compensation process, which simply removes the background common to all patterns, improves the dictionary’s ability to discriminate between diffraction patterns. However, the uncompensated dictionary will also be used since it can better discriminate anomalies that are primarily manifested in the background.

\section{Classification and Indexing}
\label{sec:classification_indexing}
The proposed correlation matching approach to classification procedes in two steps. First inner products between the observed patterns and patterns in the dictionaries $\mathcal{D}$ and $\mathcal{D}_c$ are computed. For each observed pattern we only store its $k$ closest dictionary matches, i.e., those dictionary patterns with the $k$ highest inner products with respect to the observed pattern. Then a pixel is classified as a grain interior, a grain boundary, or an anomaly using a decision tree classifier applied to the set of top pattern matches in the dictionary. The proposed correlation matching approach to indexing pixel orientation computes an estimate of the mean orientation over the top matching patterns in the dictionary. In order to account for noise and the non-euclidean nature of the Euler sphere, the mean angles are estimated using a specially adapted maximum likelihood estimator, introduced in the indexing subsection below. Pixel classification and indexing are performed independently and are discussed separately.

\subsection{Classification}
Classification of a pixel is performed by evaluating the inner products between the pixel’s uncompensated and compensated patterns and patterns stored in the dictionaries $\mathcal{D}$ and $\mathcal{D}_c$, respectively. Anomalies are detected as abnormally low average inner products between an uncompensated pixel pattern and patterns in $\mathcal{D}$. Specifically the average inner product similarity measure is defined as
\begin{equation}
\label{eq:avg_inner_product}
\bar{\rho}(i) = \frac{1}{d}\sum_{j=1}^d\frac{\bY_i^T\phi_j}{\|\bY_i\|\|\phi_j\|},
\end{equation}
where is the average of the normalized inner products between pattern $\bY_i$ at pixel $i$ and the $d$ pattern $\{\phi_j\}_{j=1}^d$ in the dictionary $\mathcal{D}$.

\begin{figure}[htb]
\centering
\centerline{\includegraphics[width=13cm]{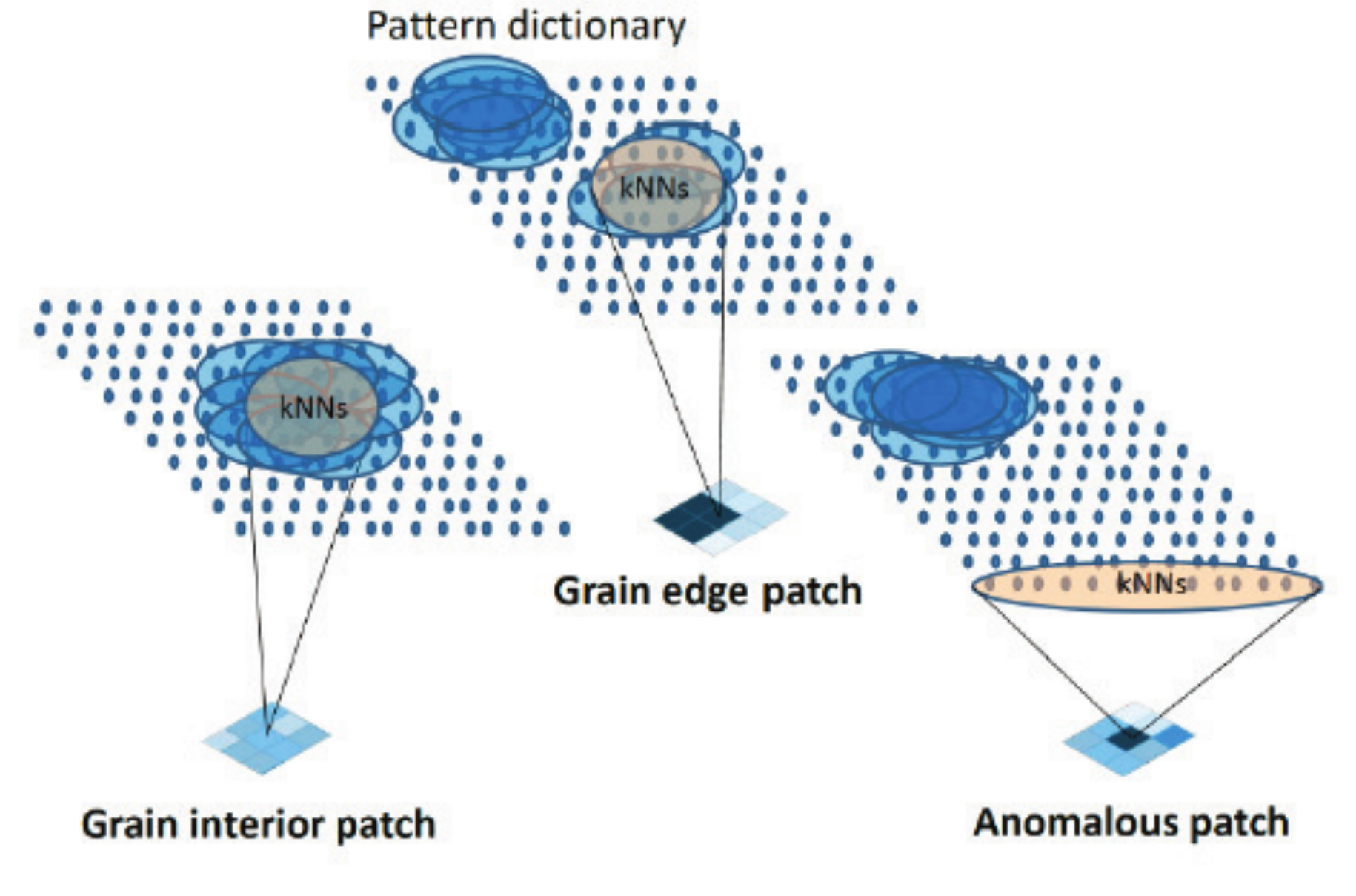}}
\caption{Illustration of a neighborhood similarity measure that quantifies the overlap between $k$-NN neighborhoods in an image patch. When the patch is inside a grain, the center of the patch will have a $k$-NN neighborhood that overlaps with the $k$-NN neighborhoods of the adjacent pixels. A patch that straddles a boundary will have the center pixel $k$-NN neighborhood overlapping with the neighborhoods of a small number of other pixels. When the patch is centered at an anomalous pixel there is little or no overlap between the $k$-NN neighborhoods of the center and adjacent pixels.}
\label{fig:interior_anomalous_figure}
\end{figure}

Boundaries between grains are detected based on the lack of homogeneity of the matches to the dictionary $\mathcal{D}_c$ of a pixel and its $8$ adjacent pixels, which we call a $3\times3$ spatial patch. Specifically, for pixel $i$ we define the neighborhood similarity measure $\rho_{\mathcal{N}_c}(i)$ as the average amount of overlap between the $k$-NN neighborhood in $\mathcal{D}_c$ of the pixel and the $k$-NN neighborhoods, with $k=40$, in $\mathcal{D}_c$ of the adjacent pixels on the patch:
\begin{equation}
\label{eq:kNN_similarity}
\rho_{\mathcal{N}_c}(i) = \frac{1}{8k}\sum_{j\in\mathcal{I}_{3\times3}(i)} card\{\mathcal{N}_{kNN}(j)\cap\mathcal{N}_{kNN}(i)\},
\end{equation}
where $\mathcal{I}_{3\times3}(i)$ are the indices of the $8$ neighbors of pixel $i$. The neighborhood similarity $\rho_{\mathcal{N}_c}(i)$ will have value close to $1$ when the image patch is located in a grain. Its value will be close to zero when the image patch is centered on an anomaly. Its value will be between zero and one when the image patch is at a grain boundary (See Fig.~\ref{fig:interior_anomalous_figure}).

\begin{figure}[htb]
\centering
\centerline{\includegraphics[width=10cm]{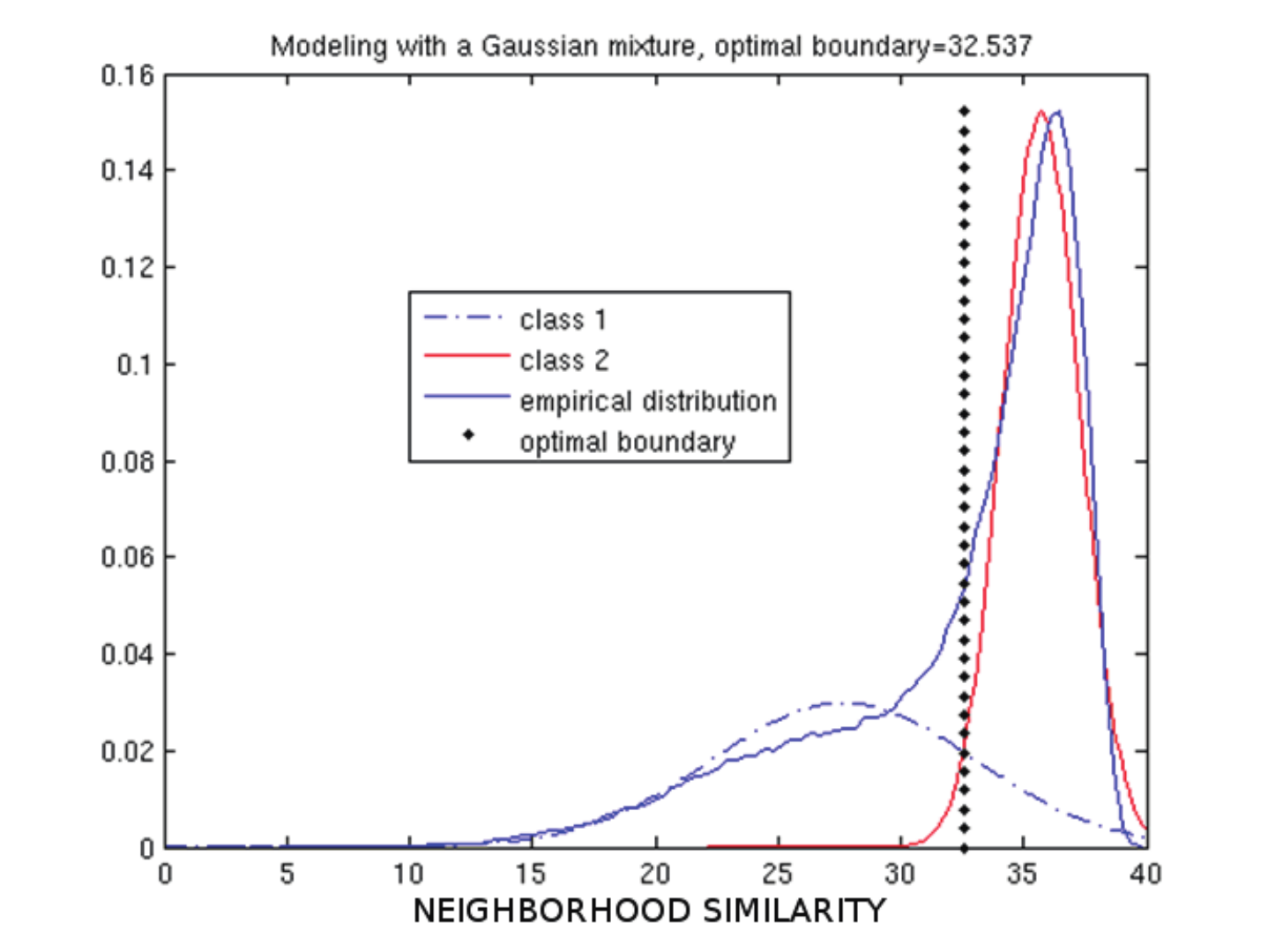}}
\caption{A two component Gaussian mixture model has a good fit to the neighborhood similarity histogram in right panel of Fig.~\ref{fig:ip_sim_hist}. The point where the two Gaussian components cross (dotted vertical line) determines the threshold for the right lower branch of the unsupervised decision tree classifier in Fig.~\ref{fig:decision_tree_exp}.}
\label{fig:similarity_mixture}
\end{figure}

\begin{figure}[htb]
\centering
\includegraphics[width=12cm]{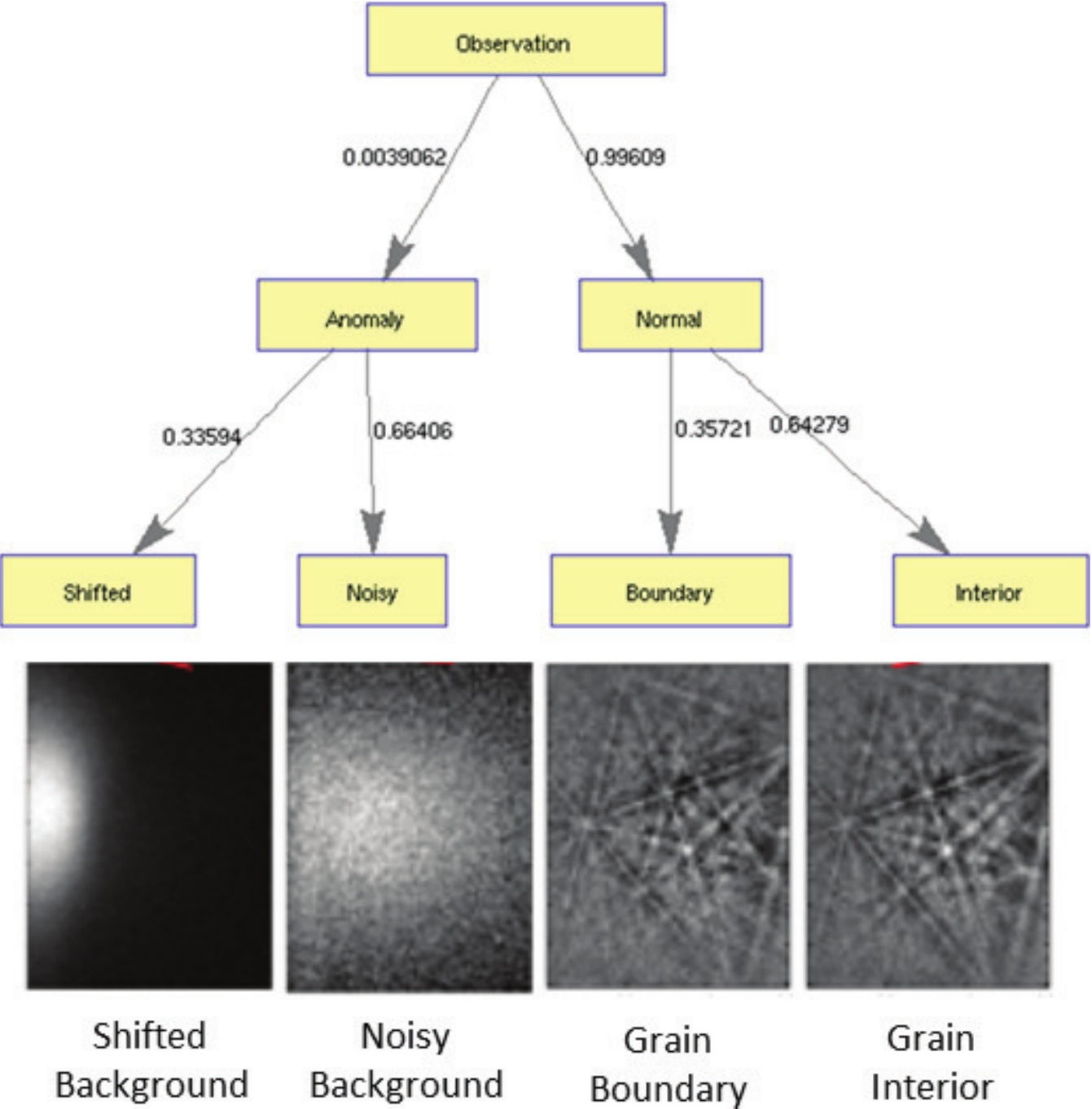}
\caption{Decision tree for clustering detected patterns on the IN100 sample with examples of patterns in each cluster below the leaf nodes at bottom. Physical locations of these patterns on the sample are shown in Fig.~\ref{fig:BSE_EBSD_sample}. The classifier uses the uncompensated pattern matches of a pixel to decide between shifted and noisy background at lower left. It uses the homogeneity of the compensated pattern matches over a $3\times3$ patch to decide between grain boundary and grain interior on the right. The number on each decision tree branch is the proportion of patterns at the parent node that were classified with label of child node.}
\label{fig:decision_tree_exp}
\end{figure}

The inner product similarities (for anomalies) and the neighborhood similarities (for grain boundaries) are used by a pattern classifier to assign each pixel to one of four classes. While many different types of unsupervised classifiers could be used, e.g., $k$-means, linear discriminant analysis (\citet{hastie_elements_2005}) or deep learning networks (\citet{hinton_fast_2006}), here we propose to cluster patterns using an unsupervised Decision Tree (DT) classifier (\citet{hastie_elements_2005}) whose classification boundaries are determined so that they separate the modes (regions of concentration) of the histograms of inner-product similarity and neighborhood similarity over the sample. The non-overlapping modes can easily be separated by thresholding of the similarity value while the others can be estimated using a mode decomposition method such as Gaussian mixture modeling, also called mixture of Gaussian (MoG) modeling (\citet{figueiredo_unsupervised_2002}). This is illustrated in Fig.~\ref{fig:similarity_mixture}. The unsupervised DT classifier is illustrated in Fig.~\ref{fig:decision_tree_exp} in Sec.~\ref{sec:results} for the IN100 sample considered. Four clusters were discovered by the model: anomalous pixels, which divided into two subclusters of shifted background and noisy background, and normal pixels, divided into grain boundary and grain interior.

Decision tree classifiers have been previously applied to many imaging applications, e.g., land cover classification in remote sensing (\citet{friedl_decision_1997}),(\citet{pal_assessment_2003}). However, there are significant differences between the proposed DT classifier and those previously applied. First, the proposed classifier is a hybrid DT that uses special features (background compensated and uncompensated patterns) and similarity measures (inner products and neighborhood intersections) specific to materials microanalysis. Second, unlike standard non-parametric DT classifiers, the proposed DT is informed by a physics model through the generated dictionary of diffraction patterns. Third, our use of unsupervised Gaussian mixture models to determine the classification thresholds means that the DT classifier threshold parameters are determined by the Gaussian mixture models and do not need to be tuned, thus eliminating the need for labeled training data and time consuming cross-validation.

\subsection{Indexing}
The proposed pixel indexing method is formulated as a statistical estimation problem. The pixel's crystal orientation is estimated via maximum likelihood under a Von Mises-Fisher mixture density model for the EUler angles of the $k$ top dictionary matches. Note that the $k$ used in the maximum likelihood model may be different from the $k$ used to compute the neighborhood similarity for the DT classifier. We motivate the Von Mises-Fisher model as follows. Recall that the dictionary is generated for a set of predetermined orientations $\btheta$. Hence using simple table lookup, the indices of dictionary patterns found in the $k$-NN neighborhood of a pixel can be mapped to a set of $k$ orientations $\{\btheta_j\}_{j=1}^k$. If the pixel is in a grain then these orientation will be clustered around a true crystal orientation, that we call $\btheta$, at the pixel location. We extract a maximum likelihood estimate of this orientation using a statistical model for the variation of $\btheta_j$'s.

We assume that the $k$ best matching orientations $\{\btheta_j\}_{j=1}^k$ of a pixel form a random sample from an underlying marginal density $f(\dot;\btheta)$, supported on the orientation sphere. Then the maximum likelihood estimator of $\btheta$ is 
\begin{equation}
\label{eq:theta_ML_formula}
\hat{\btheta} = \argmax_\btheta \prod_{j=1}^kf(\btheta_j;\btheta).
\end{equation}

As is customary in the theory of directional statistics (\citet{mardia_directional_1999}) we parameterize the density of $3$-dimensional orientations by their equivalent $4$-dimensional unit length quaternions $\{\bq_j\}_{j=1}^k$ ($\|\bq_j\|_2=1$). Due to crystal symmetry, there are many ($M$) orientations that are equivalent to each other, i.e., the Euler angle representation is not unique. For any quaternion $\bq$, the set of symmetry equivalent quaternions can be represented as $\{\bP_m\bq\}_{m=0}^{2M-1}$ where $\bP_m$ is a $4\times4$ quaternion (rotation) matrix and $\bP_0$ is the identity matrix. The matrices $\{\bP_m\}_{m=0}^{2M-1}$ are symmetric and constitute an abelian group of actions on the $3$-dimensional sphere. For example, Nickel has face-centered cubic symmetry, hence there are $M=24$ symmetry-equivalent orientations. Since the quaternion representation of orientations associates two quaternions ($\bq$ and $-\bq$) to each orientation, we need $48$ matrices $\{\bP_m\}_{m=0}^{47}$ to establish a $4$D representation of the $m\bar{3}m$ point group (\citet{de_graef_structure_2007}).

The proposed model for orientations is based on a generalization of the Von Mises-Fisher density (\citet{mardia_directional_1999}) to group structured domains on the sphere. The standard Von Mises-Fisher density over the $3$-dimensional sphere with location parameter $\bmu$ and precision parameter $\kappa$ is defined as
\begin{equation}
\label{eq:VMF_density}
f_{VMF}(\bx;\bmu,\kappa) = c_4(\kappa)\exp{(\kappa\bmu^T\bx)}, \|\bx\|_2=1
\end{equation}
where $c_p(\kappa)=\frac{\kappa^{p/2-1}}{(2\pi)^{p/2}I_{p/2-1}(\kappa)}$ and $I_\nu$ is the modified Bessel function of the first kind. Here $\|\bmu\|_2=1$ and $\kappa>0$ control the location of the mode (maximum) and spread of the density over the sphere, respectively. The natural generalization to the group structured domain is the periodic mixture density 
\begin{equation}
\label{eq:VMF_density_mixture}
f(\bx;\bmu,\kappa) = \frac{1}{2M}\sum_{m=0}^{2M-1} f_{VMF}(\bx; \bQ_m\bmu,\kappa).
\end{equation}
This Von Mises-FIsher mixture (VMFm) density contains $2M$ replicates of the Von Mises-Fisher density over the sphere centered at all the symmetry-equivalent values of the location parameters $\bmu$.

Substitution of (\ref{eq:VMF_density_mixture}) into (\ref{eq:theta_ML_formula}) and use of the well-known invariance property of maximum likelihood estimation (\citet{lehmann_theory_1998}) gives a form for the maximum likelihood estimator of grain orientation $\hat{\btheta}$ in terms of the joint maximum likelihood estimators $\hat{\bmu}$ and $\hat{\kappa}$:
\begin{equation}
\label{eq:VMF_density_completedata}
\{\hat{\bmu},\hat{\kappa}\} = \argmax_{\bmu,\kappa} \prod_{j=1}^k\sum_{m=0}^{2M-1}\gamma_mf_{VMF}(\bq_j;\bQ_m\bmu,\kappa).
\end{equation}
where $\gamma_m=(2M)^{-1}$. Even though this maximization problem appears daunting, it can be iteratively and efficiently computed by applying the well known expectation-maximization (EM) procedure for constrained parameter estimation in mixture models (\citet{mclachlan_finite_2004}). For a full account of this procedure we refer the interested reader to~\citet{chen_parameter_2015}. Note that $1/\hat{\kappa}$ gives an empirical estimate of the degree of spread of the density about the orientation estimate $\hat{\bmu}$ Thus, $\hat{\kappa}$ is a measure of confidence of this estimate.

\section{Generation of the Dictionary}
\label{sec:Dict_generation}
The dictionary approach requires a uniform sampling of orientation space $SO(3)$. Several sampling schemes are available in the literature; among the most popular schemes are a deterministic sampling method based on the Hopf fibration (\citet{yershova_deterministic_2004},\citet{yershova_generating_2009}) and the HEALPix framework (Hierarchical Equal Area isoLatitude Pixelization) (\citet{gorski_healpix:_2005}). Neither of these approaches is easily adaptable for integration with crystallographic symmetry. Instead, we employ a recently developed strategy that starts from a simple $3$D cubic grid which is mapped uniformly onto $SO(3)$ (\citet{rosca_new_2014}). This cubochoric mapping is uniform, refinable, and isolatitudinal, and consists of three steps:
\begin{enumerate}
\item a uniform cubic grid of $(2N+1)^3$ grid points is generated inside a cube of edge length $a=\pi^{2/3}$;
\item the cube is divided into six pyramids with apex at the cube center and the cube faces as base, and each pyramid is mapped uniformly onto a sextant of a ball, using a generalization of the mapping of a square onto a curved square (\citet{rosca_uniform_2011});
\item all points inside the ball are then transformed, using a generalized inverse equal-area Lambert mapping, to the unit quaternion Northern hemi-sphere, which is isomorphous with $SO(3)$.
\end{enumerate}

From the quaternion representation one can readily derive other orientation parameterizations; the Rodrigues parameterization is most suitable for the determination of the orientations that belong to the fundamental zone (FZ) for a given crystal symmetry, because the boundaries of the FZ are planar. The more conventional Euler angle representation typically has curved surfaces as the boundaries of the FZ, so that Euler angles are less useful for uniform sampling approaches. It should be noted that lower crystal symmetry implies a larger dictionary, since the fundamental zone size increases with a reduction in symmetry; acceleration of the dot product calculations by means of a GPU (graphical processing unit) is a topic of ongoing research. The use of the Rodrigues representation to determine the dictionary elements has computational advantages, but care must be taken in the case of crystal symmetries with a single diad axis, for which the Rodrigues fundamental zone becomes infinite in the direction normal to the axis. Our approach still produces a uniform sampling of orientation space, although all rotations by an angle of $180$\textdegree\  are represented by points at infinity (which correspond to points on the outer cube surface in the cubochoric representation).

\begin{figure}[htb]
\centering
\includegraphics[width=14cm]{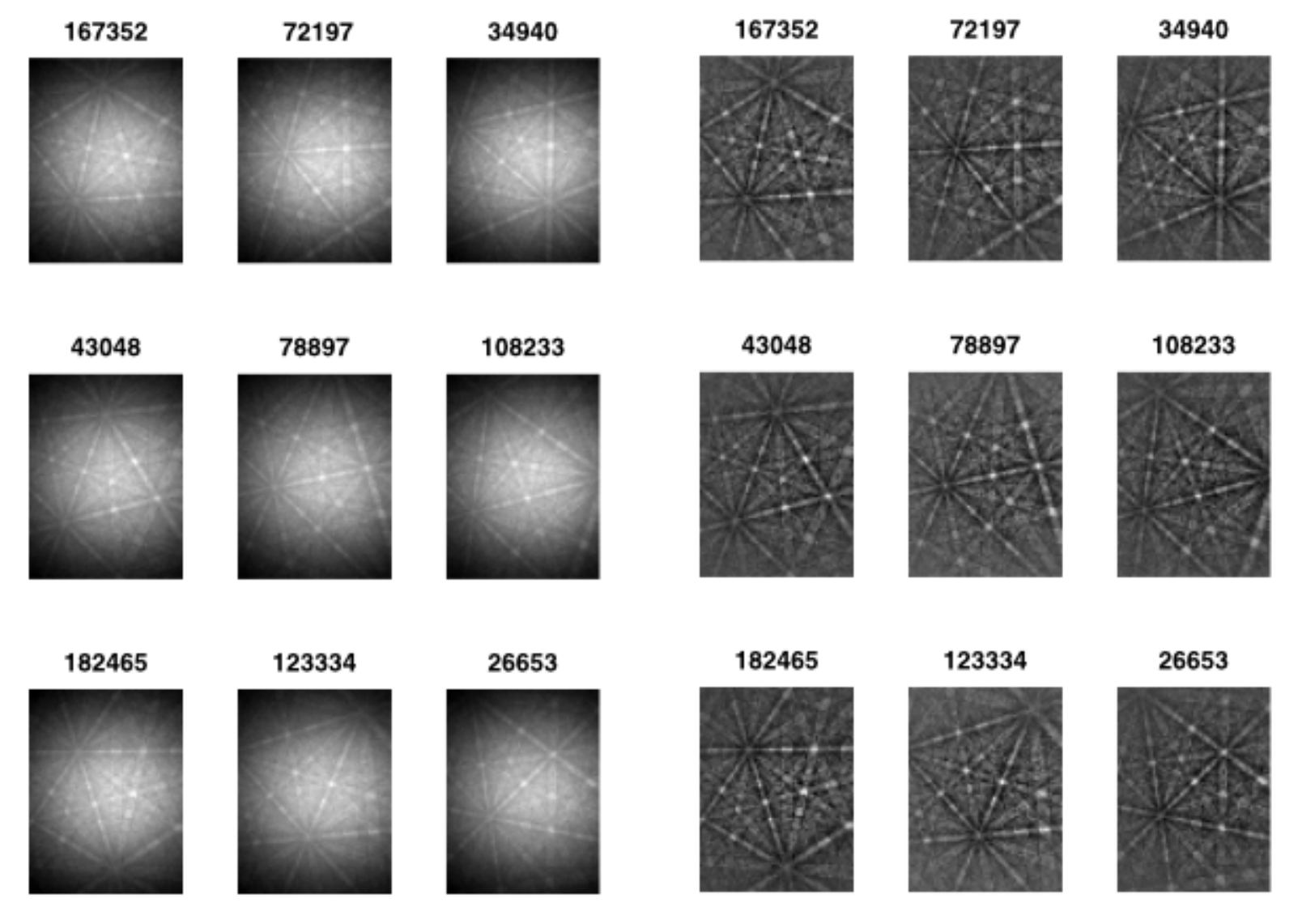}
\caption{A random subset of the $333226$ elements in the dictionary generated for the IN100 sample. Shown are $9$ representative patterns, each $60×80$ pixels, in the uncompensated (Left) and compensated (Right) versions of the dictionary.}
\label{fig:patterns_examples}
\end{figure}

\begin{figure}[htb]
\centering
\includegraphics[width=14cm]{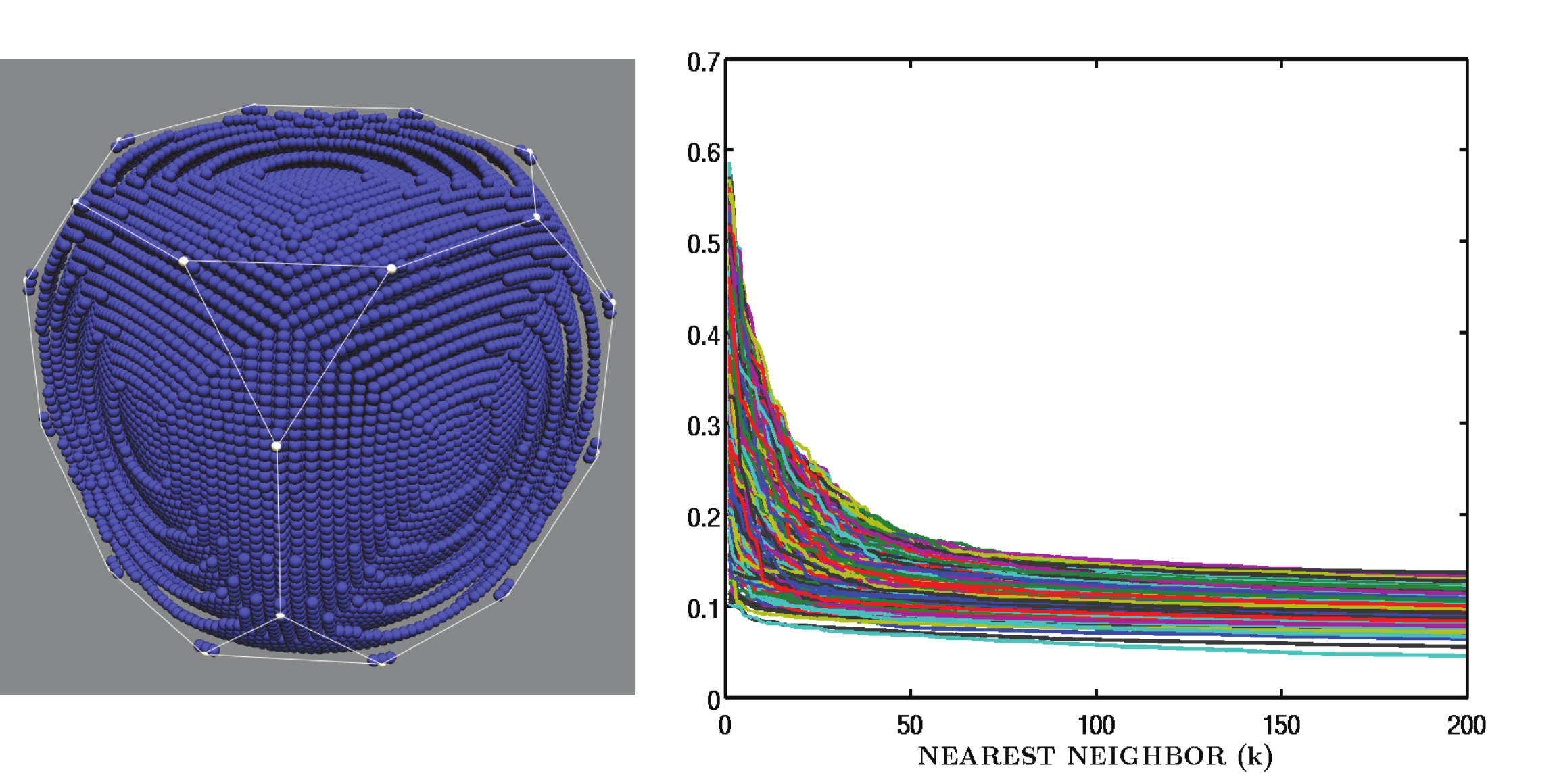}
\caption{Left: The sampling pattern (at $1/8$ density) of dictionary Rodrigues vectors in the fundamental zone (solid lines) of the cubic symmetry point group $m\bar{3}m$. Right: Graph of the top $200$ normalized inner products between the entire compensated IN100 dictionary and a randomly selected set of $4000$ reference elements in the IN100 dictionary. For each of the reference elements the top $200$ inner products have been rank ordered in decreasing order and plotted. A knee occurs in vicinity of $k = 40$ for which the normalized inner product drops by at least $1/3$ of the maximum value.}
\label{fig:samplinggrid_ip}
\end{figure}

The dictionary used for the remainder of this paper was generated by setting $N=100$, and keeping only those orientations for which the corresponding Rodrigues vector lies inside the fundamental zone for the octahedral $m\bar{3}m$ point group. The patterns in the dictionary were down sampled to $60$ by $80$ pixel images. This results in a dictionary $\mathcal{D}$ with $d=333226$ elements of dimension $4800$. A representative (random) selection of $9$ dictionary elements in $\mathcal{D}$ and $\mathcal{D}_c$ is shown in Fig.~\ref{fig:patterns_examples}. The left panel of Fig.~\ref{fig:samplinggrid_ip} shows how the sampling points are distributed inside the octahedral Rodrigues FZ. The right panel of the figure shows the rate of drop-off of the top $200$ inner products in the compensated dictionary $\mathcal{D}_c$ for $4000$ randomly selected reference elements. This decay rate is used to select the number of nearest neighbors ($k=40$) used for the classifier described in Sec.~\ref{sec:classification_indexing} and implemented in Sec.~\ref{sec:results}.

\section{Computational Considerations}
\label{sec:computation}
The online dictionary matching algorithm requires inner product evaluation between the measured sample diffraction patterns and the dictionary diffraction patterns. Let $d$ denote the number of patterns in the dictionary (dictionary size), $L$ denote the number of pixels on the photodetector (pattern size), and $n$ denote the number of pixels on the sample (sample size). The time complexity of calculating the mean inner product over the entire measured sample is $O(L(d+n))$.

For the indexing method, to determine the $k$-NN dictionary patterns for a given pixel the $k$ largest inner products need to be determined from the set of all inner products between the pixel and patterns in the dictionary. The time complexity of the whole process is $O(Ldn)$, assuming $k\ll d$. The computation time and space grow rapidly when the dictionary size and sample size become large.

Computational challenges can be addressed in several ways. The simplest approach is to use parallelization and distributed computation. All of the algorithms introduced here can be parallelized over the spatial domain since they involve local operations. For example, ML orientation estimation is applied independently to each pixel and DT classification is applied to each spatial patch in the sample. To speed up the $k$-NN dictionary search one can use methods from information retrieval such as dictionary caching and KD trees to accelerate the inner product evaluation and ranking process. These methods rely on the similarity of diffraction patterns over the $k$-NN neighborhood. However, as they rely on approximation, these methods may also introduce indexing errors. A study of these and other computational trade-offs is important but is outside the scope of this paper.

\section{Experimental Methods}
\label{sec:exp_methods}
To test the dictionary approach against an experimental data set, a polycrystalline IN100 Nickel-based super-alloy sample was selected. The sample was polished using a multi-platen Robomet.3D, using a grit of $1$ micron diamond slurry on a TexMet cloth and finished with a $40$ nm colloidal silica slurry on a ChemoMet cloth. Between polishing steps, a water clean was used on a ChemoMet cloth.

A backscattered electron (BSE) image was recorded using a Tescan Vega $3$ XMH scanning electron microscope outfitted with a $LaB_6$ filament. An EBSD map was obtained with the same SEM and a Bruker e-Flash1000 system. A tilt angle of $70$\textdegree, voltage of $30$kV, working distance of $15$mm, and emission current of about $1$nA were used to collect the EBSD map. The spatial resolution in both $x$ and $y$ directions was $297.7$nm, and a Kikuchi pattern of $80\times 60$ pixels was acquired and stored at every point in a $512\times 382$ map.

\section{Results}
\label{sec:results}
A dictionary was designed as described in Sec.~\ref{sec:Dict_generation} for the octahedral $m\bar{3}m$ crystal symmetry group to match the known characteristics of the IN100 sample and the SEM system, as described in Sec.~\ref{sec:exp_methods}. Dictionary inner-products and $k$-NN neighborhoods were computed from the detected pattern at each scan location.

\begin{figure}[htb]
  \centering
  \includegraphics[width=14cm]{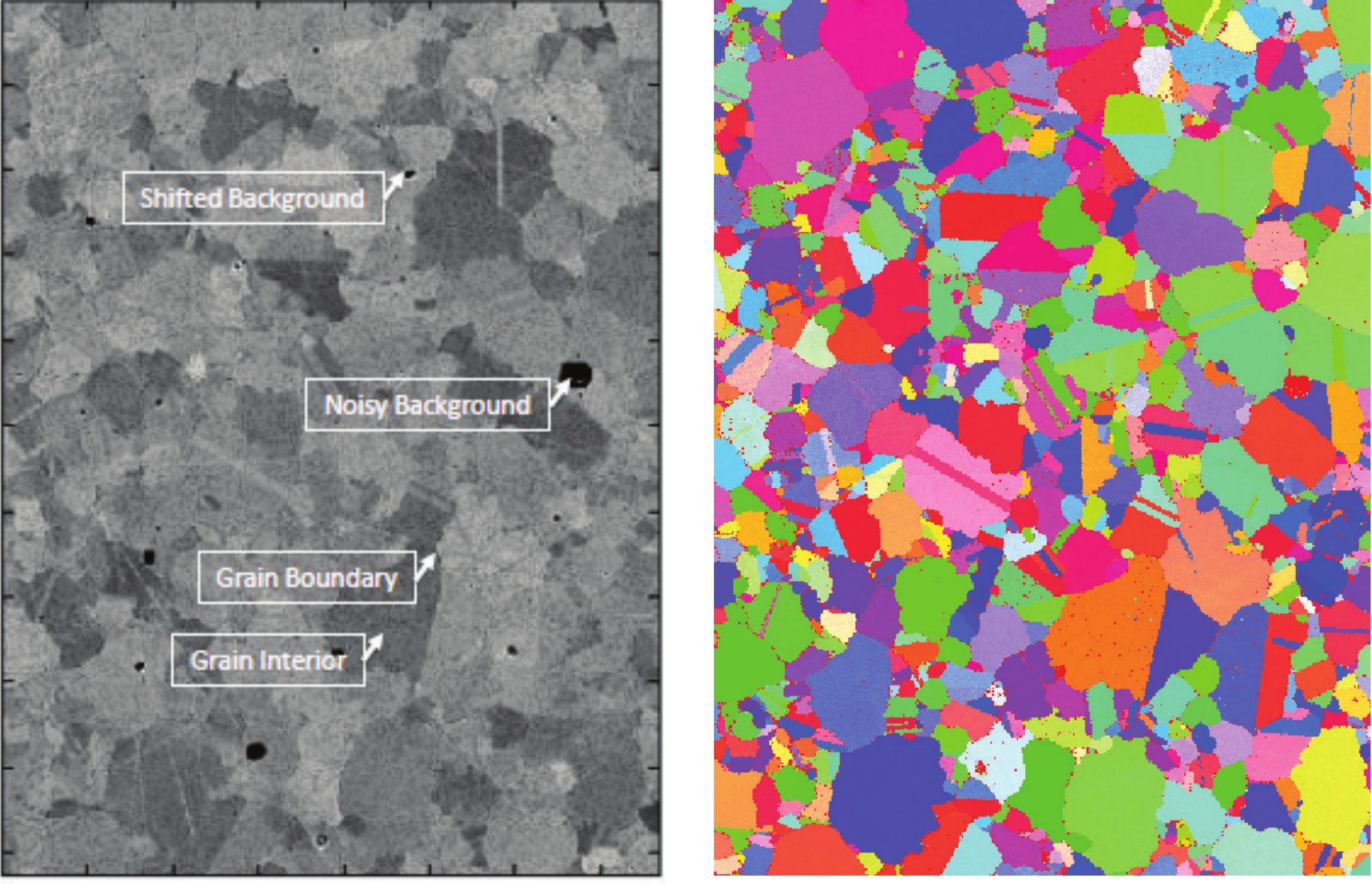}
\caption{Raw SE and EBSD images of IN100 sample generated by the Tescan Vega SEM with native OEM software. Left: SE image of the IN100 sample showing physical locations of the four patterns shown at bottom of DT classifier in Fig.~\ref{fig:decision_tree_exp}. The inner-product histograms for the diffraction patterns at these locations are shown in Fig.~\ref{fig:avg_inner_product}. Right: IPF colored EBSD pixel orientation image.}
\label{fig:BSE_EBSD_sample}
\end{figure}

\begin{figure}[htb]
\centering
\includegraphics[width=10cm]{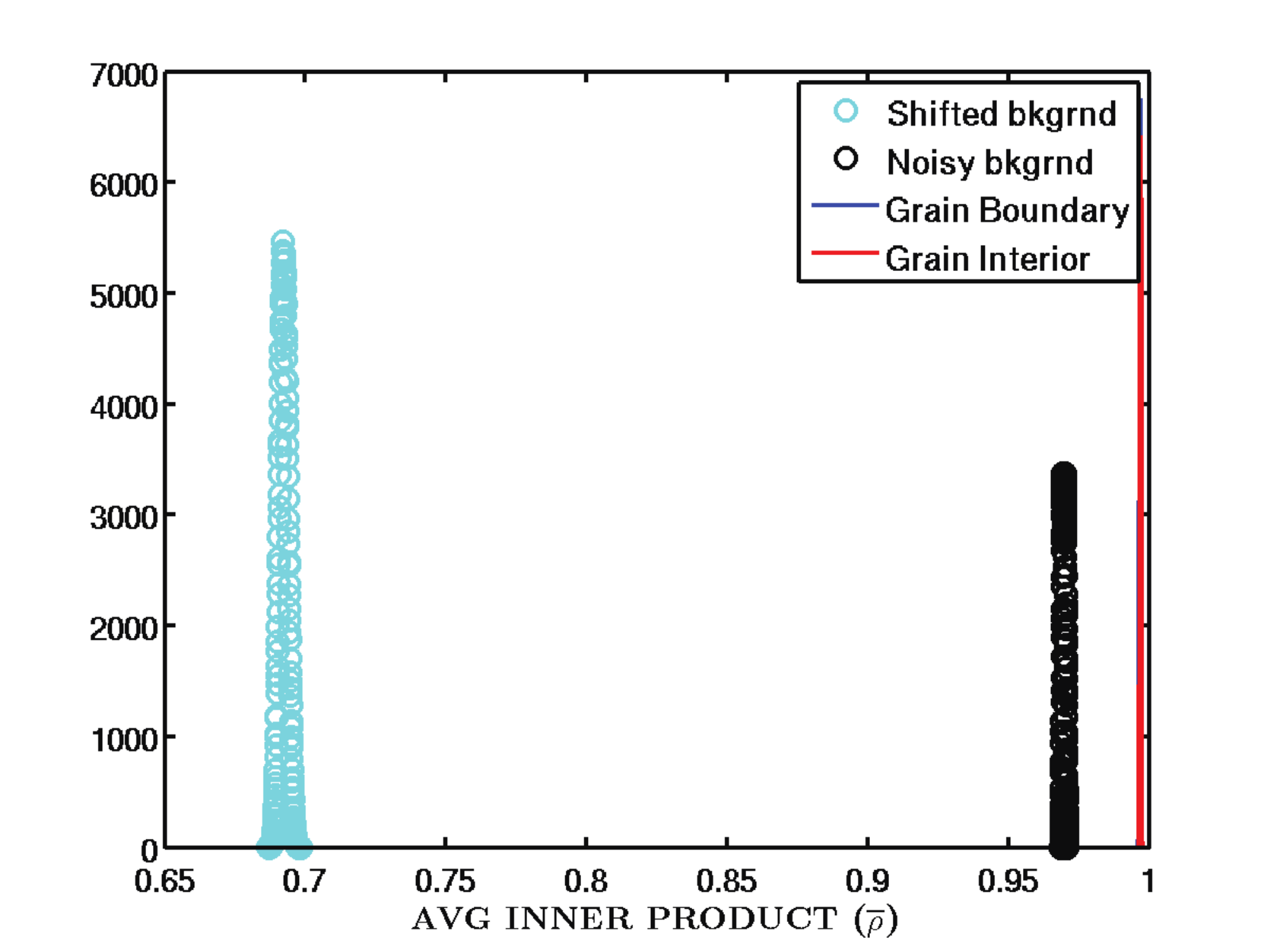}
\caption{Histograms of the inner products between patterns in the dictionary and the patterns of the four EBSD scan locations (pixels) shown in Fig.~\ref{fig:BSE_EBSD_sample}. The histograms for the “shifted background” and the “noisy background” are well separated from each other and from the histograms for the “grain boundary” and “grain interior” pixels in Fig.~\ref{fig:BSE_EBSD_sample}. These latter two histograms are very concentrated near $1$ and overlap each other (not distinguishable at this scale).}
\label{fig:avg_inner_product}
\end{figure}

We indicate four different scan locations (pixels) with qualitatively different patterns in Fig.~\ref{fig:BSE_EBSD_sample}. These locations are representative of the four different clusters of pattern, described below (see Fig.~\ref{fig:decision_tree_exp}), that were discovered by the unsupervised DT classifier. In Fig.~\ref{fig:avg_inner_product} the histograms of the dictionary inner-products are shown for each of these pixels. The "shifted background" and "noisy background" pixels have inner-products that are well separated from each other in addition to being separated from the inner-products of the "grain interior" and "grain boundary" pixels. Thus one might expect that the group of "anomalous" pixels, represented by the former two, could be easily separated from the group of "normal" pixels, represented by the latter two, using any reasonable clustering technique based on the inner-products. On the other hand, the inner-product histograms for the grain interior pixel and the grain boundary pixel are overlapping. This overlap makes it difficult to distinguish these two types of pixels and justifies the need for the more sensitive neighborhood similarity measure that is better able to separate them.

\begin{figure}[htb]
  \centering
  \includegraphics[width=15cm]{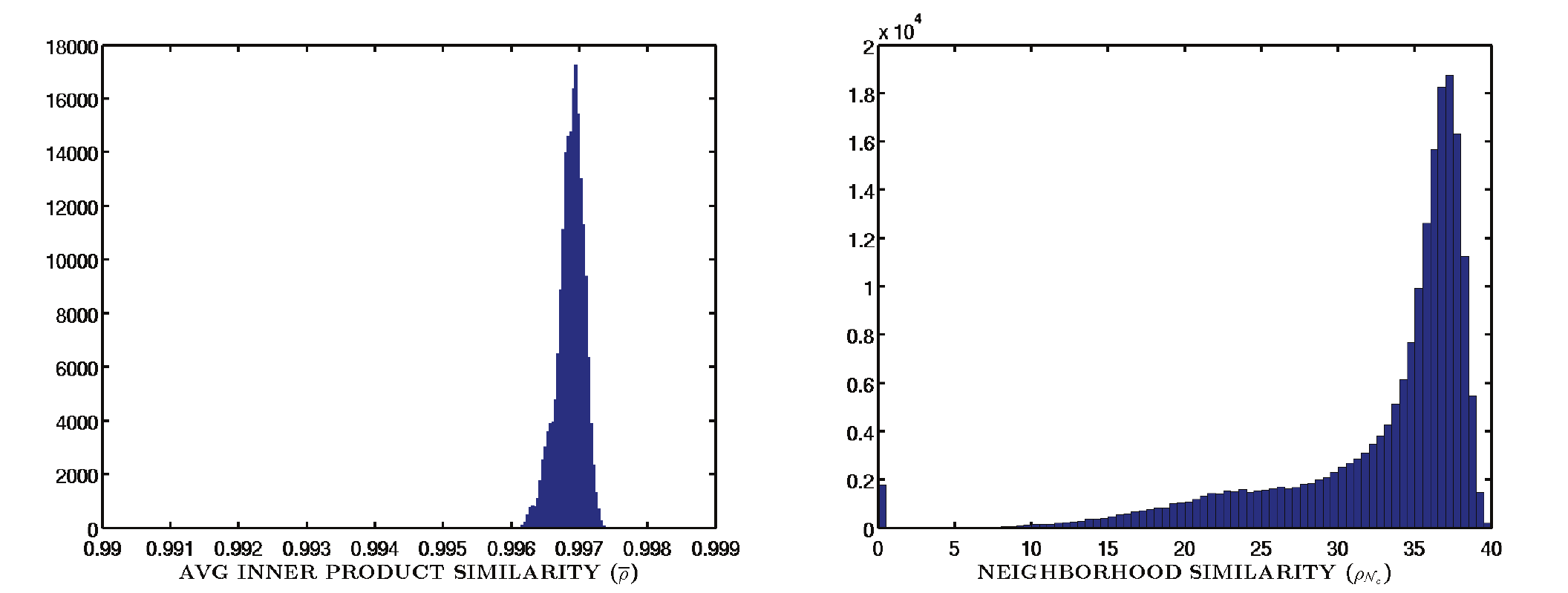}
\caption{Left: histogram of normalized inner products between detected patterns on the sample and dictionary patterns restricted to the range $\bar{\rho}=[0.99,0.999]$ to reveal the modes associated with grain interior and grain boundary patterns. Two other modes (not shown) are located near $\bar{\rho}=0.7$ and $\bar{\rho}=0.97$ corresponding to background shift and noisy background pixels, respectively (see Fig.~\ref{fig:avg_inner_product}). Right: histogram of neighborhood similarity measures between dictionary neighborhoods over a $3\times3$ patch centered at each pixel in the sample for neighborhood size $k=40$.}
\label{fig:ip_sim_hist}
\end{figure}

Figure~\ref{fig:ip_sim_hist} shows the full-sample histograms of inner-products $\bar{\rho}$ with respect to $\mathcal{D}$ and neighborhood similarities $\rho_{\mathcal{N}_c}$ with respect to $\mathcal{D}_c$, respectively, over all pixels and over all patterns in the dictionaries. The left panel of Fig.~\ref{fig:ip_sim_hist} can be interpreted as the addition of all other inner-product histogram to the four histograms shown in Fig.~\ref{fig:avg_inner_product}. Similarly to Fig.~\ref{fig:avg_inner_product}, the full-sample inner-product histogram exhibits three well separated modes, which confirms that anomalous pixels can easily be separated from the normal pixels on the basis of thresholding each pixel's average inner product measure. The three modes correspond to anomalous pixels with inner-products clustered around $0.7$ and $0.97$ (not visible in the range of $\bar{\rho}$ plotted) and normal pixels with inner-products clustered around $0.997$.

The right panel of Fig.~\ref{fig:ip_sim_hist} shows the histogram of all neighborhood similarities computed with neighborhood size $k=40$. The latter histogram is bimodal and asymmetric about its mean. The higher mode located near $37$ corresponds to pixels whose $k$-NN neighborhood in $\mathcal{D}_c$ has high overlap with the $k$-NN neighborhoods of its $8$ adjacent pixels. Such pixels are likely to be interior to a grain. The lower mode located near $26$ corresponds to patches of pixels near grain boundaries, patches that have less similar $k$-NN neighborhoods than in-grain pixels. To separate these tow modes, we fitted a two component mixture-of-Gaussian model to the histogram in the middle panel using the MoG EM algorithm (\citet{mclachlan_finite_2004}). The result of this fit is shown in Fig.~\ref{fig:similarity_mixture}. The point of intersection of each of the fitted Gaussian densities (shown in the Figure by vertical dotted line) is used as the DT classification threshold for discriminating between grain boundaries and grain interiors.

Figure~\ref{fig:decision_tree_exp} shows the unsupervised DT classifier used to cluster observed patterns. The lower four nodes are leaf nodes while the upper three nodes are decision nodes for which thresholds are used to assign labels. These thresholds were determined from the observed histograms shown in Fig.~\ref{fig:ip_sim_hist} as described above. The DT classifies each pixel on the sample based on the pattern matches in the dictionaries. The top node, labeled "observation" classifies pixel as a "anomaly" or as "normal" by thresholding the average inner-product $\bar{\rho}$ between the pattern at the center pixel of the patch and the patterns of the dictionary. Any threshold between $\bar{\rho}=0.97$ and $\bar{\rho}=0.99$ would separate the normal pair (grain interior, grain boundary) from the anomalous pair (noisy background, shifted background) and we selected the midpoint. The anomalous patterns are further subclassified on the left branch of the DT by applying a threshold between $0.7$ and $0.95$ to $\bar{\rho}$. The DT classifies normal pixels as either grain-interior or grain-boundary pixels by applying the threshold $32/40$ to the neighborhood similarities $\rho_{\mathcal{N}_c}$. Representative patterns are shown at the bottom of Fig.~\ref{fig:decision_tree_exp} that have been identified as belonging to each of the respective four clusters.

\begin{figure}[htb]
  \centering
  \includegraphics[width=13cm]{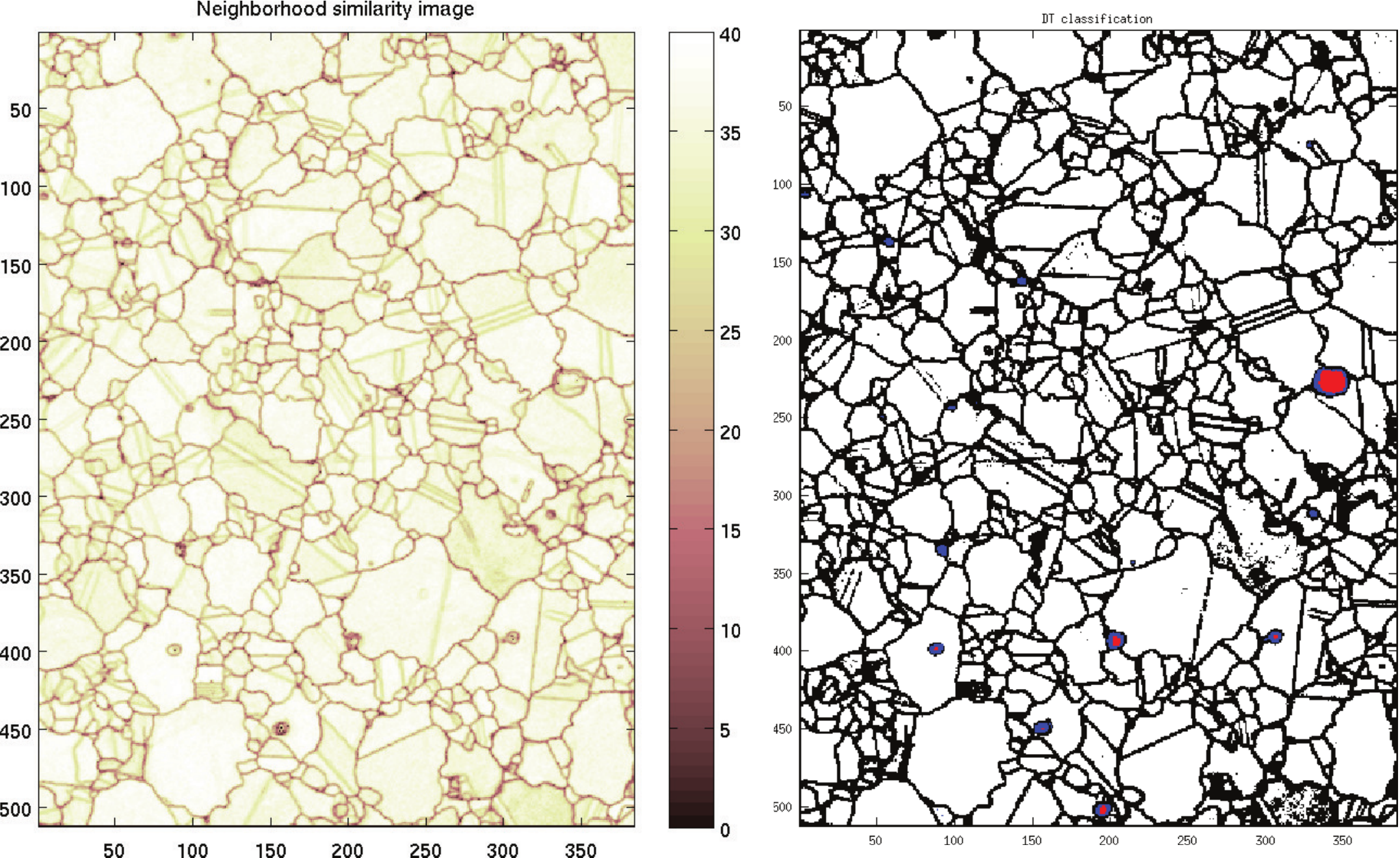}
\caption{Left: An image rendering of the (un-normalized) neighborhood similarity measure ($k = 40$ nearest neighbors in dictionary) used in the right branch of the DT classifier in Fig.~\ref{fig:decision_tree_exp}. Right: A map of the pattern classes in the IN100 sample as determined by the DT classifier in Fig.~\ref{fig:decision_tree_exp}. The colors encode the four classes as follows: white=grain interior, black=grain boundary, red=noisy background, and blue=shifted background. Note that the black boundaries effectively segment the sample according to crystal orientation.}
\label{fig:simmap_classification_img}
\end{figure}

\begin{figure}[htb]
  \centering
  \includegraphics[width=13cm]{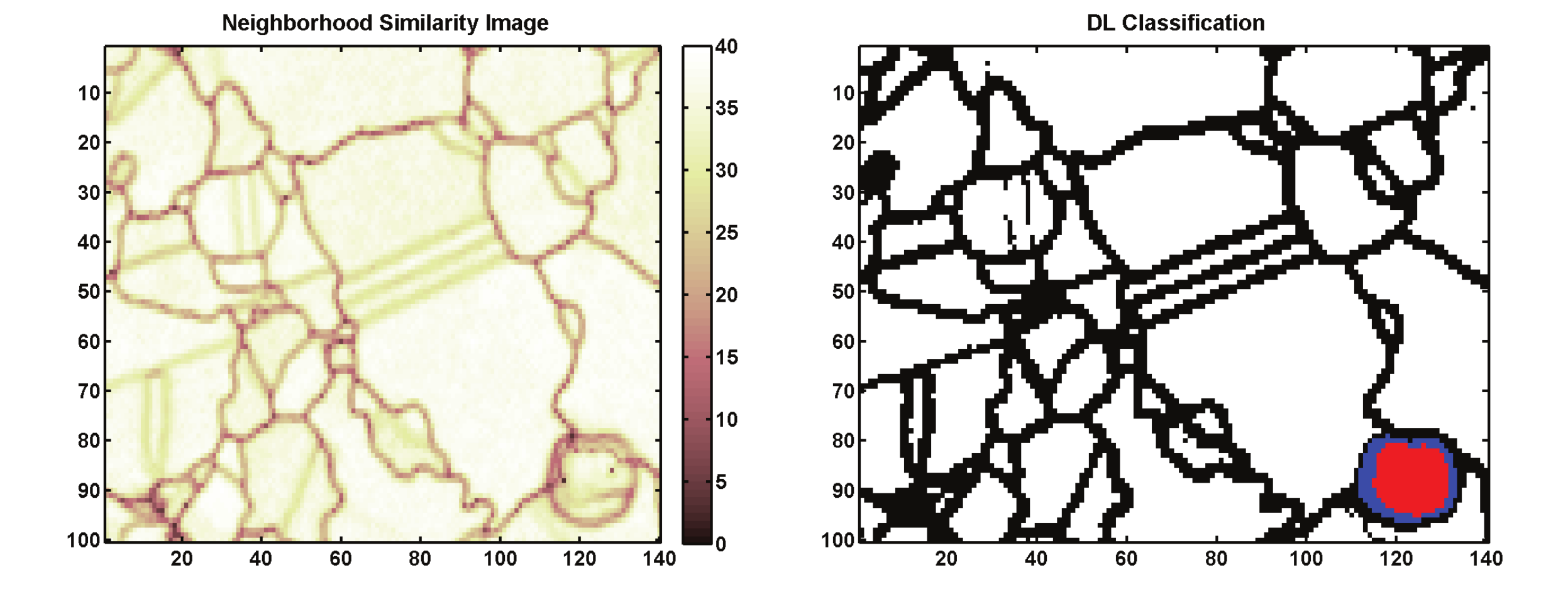}
\caption{Blowup of a small region right of center in each of the images of Fig.~\ref{fig:simmap_classification_img}.}
\label{fig:simmap_classification_subimg}
\end{figure}

Figure~\ref{fig:simmap_classification_img} shows the pixel neighborhood similarities (left panel) and the pixel classifications (right panel) as images as determined by the unsupervised DT classifier. In the classification image the blue/red regions and black regions respectively correspond to pixels classified as anomalies and boundaries. These class labels can be used for segmentation of the grains and identification of the anomalies. A blowup of these images in Fig.~\ref{fig:simmap_classification_img} is shown in Fig.~\ref{fig:simmap_classification_subimg} for a small region.

\begin{figure}[htb]
\centering
\includegraphics[width=13cm]{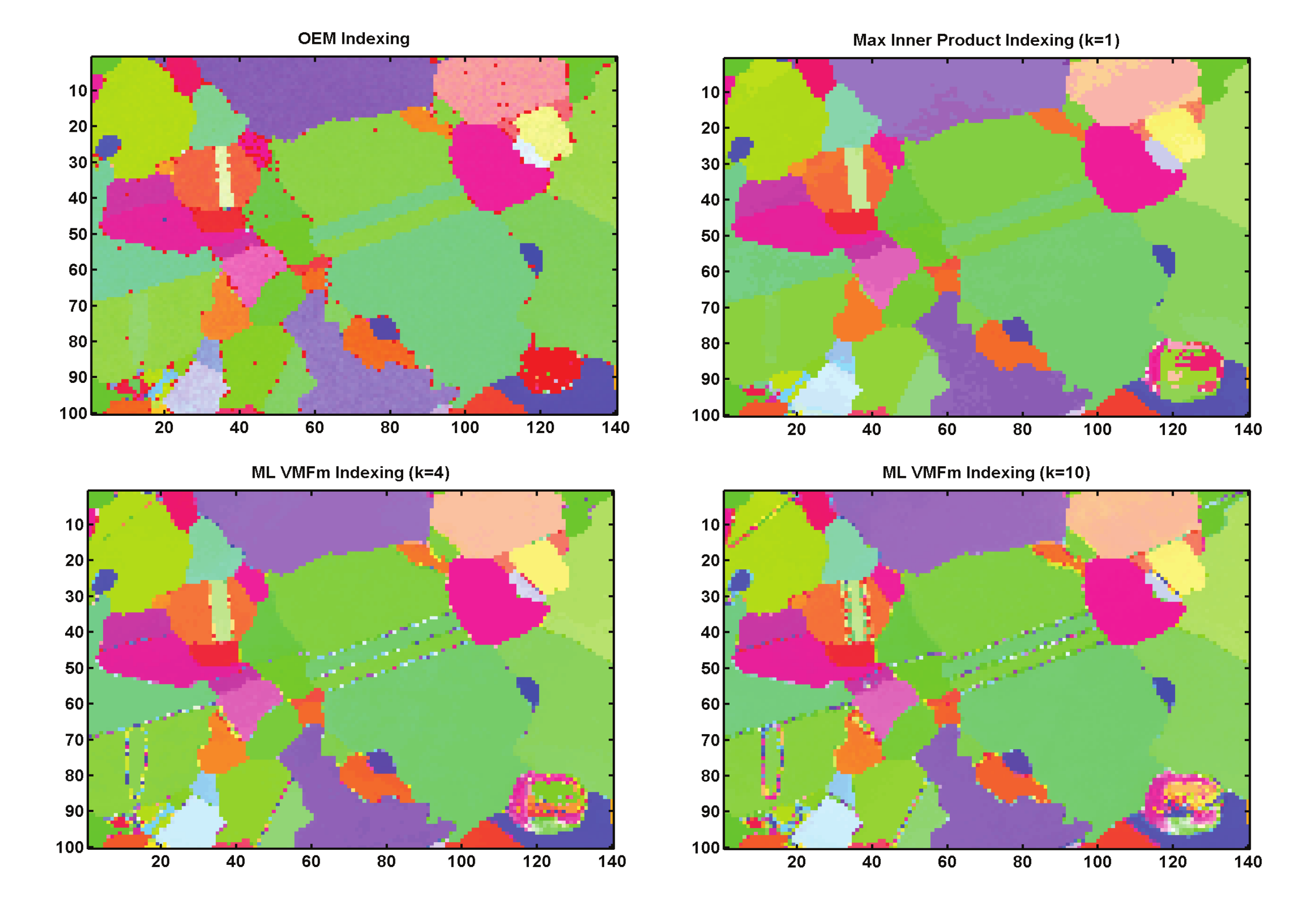}
\caption{Comparison of orientation indexing. Top left: IPF images generated by OEM software. Top right: IPF image obtained by rendering the top matching patterns in the dictionary (this is identical to the ML estimator of the orientation using VMFm model with $k = 1$). Bottom left: Image of ML estimates of orientation using VMFm model on the orientations of the $k = 4$ top dictionary matches. Bottom right: Same as bottom left except that $k = 10$. Note that the OEM image has many spurious orientation estimates within grains unlike the other dictionary based methods. Note also that the ML orientation estimates produce smoother in-grain orientations. The $k = 4$ and $k = 10$ ML orientation estimates have low confidence (high variance) at some locations on grain boundaries and in anomalous region at bottom right. This low confidence is quantified by the ML estimator of the scale parameter $\kappa$ of the VMFm model, shown in Fig.~\ref{fig:indexing_uncertainty} for $k = 4$ and $k = 10$.}
\label{fig:indexing_IPF}
\end{figure}

\begin{figure}[htb]
\centering
\includegraphics[width=14cm]{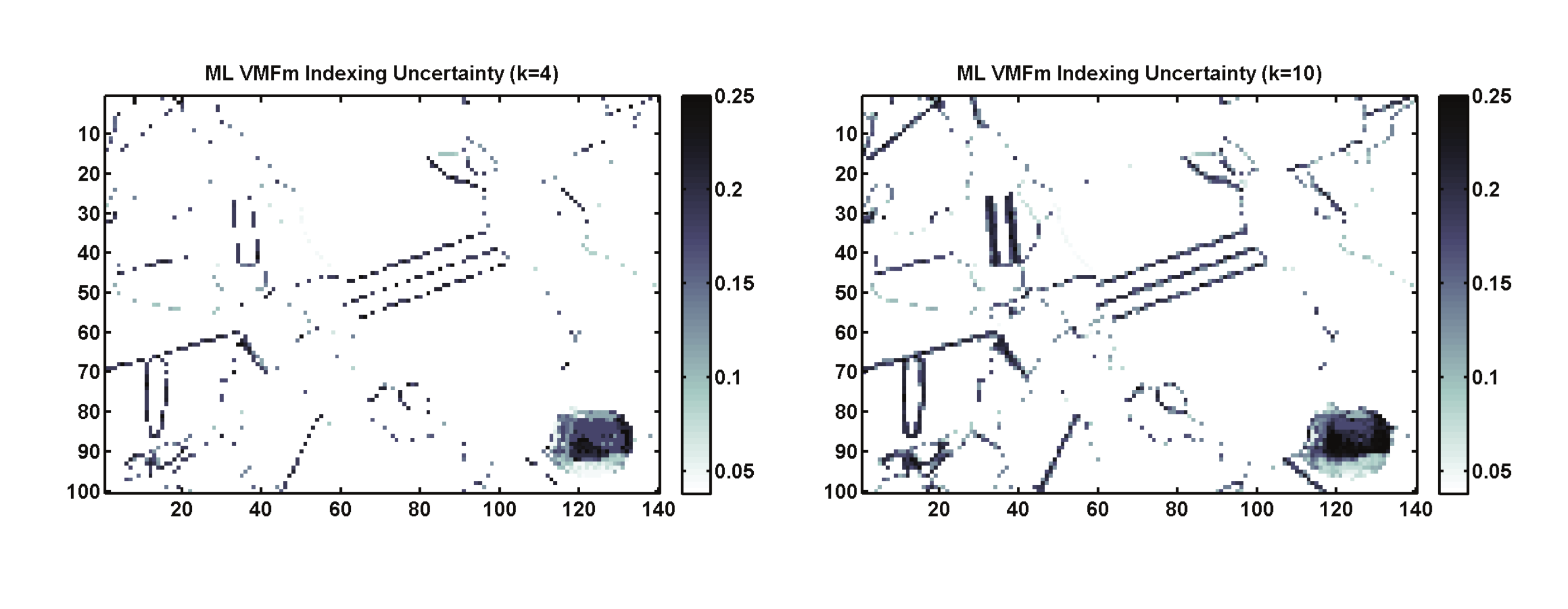}
\caption{Images of the ML estimator of the orientation standard deviation (in degrees) obtained by ML estimation of the scale parameter $\kappa$ of the VMFm model corresponding to the bottom two sub-figures of Fig.~\ref{fig:indexing_IPF}. The angular standard deviation ranges from $0.05$ degrees to $0.5$ degrees but those values above $0.25$ have been hard-limited for ease of visualization (only $1\%$ of all values are above $0.25$ degrees). Note that the areas of least confidence are in the vicinity of boundaries and anomalies. The highest standard deviations occur at pixels that straddle boundaries between grains having the highest misorientation.}
\label{fig:indexing_uncertainty}
\end{figure}

Next we illustrate the use of the dictionary for estimation of the Euler angles in the sample. For the same subregion as in Fig.~\ref{fig:simmap_classification_subimg}, Fig.~\ref{fig:indexing_IPF} illustrates the OEM (original equipment manufacturer) orientation estimates (top left image), an estimate equal to the orientation of the element of the dictionary having largest normalized inner product with the pixel pattern (top right image), the proposed ML estimator $\hat{\bmu}$ of the orientation (VMFm location parameter) computed from the top $k=4$ matches in the dictionary (bottom left image), the proposed ML VMFm estimator $\hat{\bmu}$ computed with $k=10$ (bottom right image). The ML estimates $\hat{\kappa}$ of the VMFm scale parameters were computed for each pixel. This $\hat{\kappa}$ parameter is inversely proportional to the width of the estimated VMFm model over the $3$ dimensional quaternion sphere. Figure~\ref{fig:indexing_uncertainty} shows images of these ML estimates translated into angular uncertainty (in degrees) by using the transformation $\Delta\theta=\arccos(1-1/\kappa)\frac{180}{\pi}$, which is the $1/e$-width of the VMFm distribution in any fundamental zone. Note that the OEM image in Fig.~\ref{fig:indexing_IPF} has many spurious orientation estimates within grains unlike the proposed dictionary based methods. Note also that the ML orientation estimates produce smoother in-grain orientations. The $k=4$ and $k=10$ ML orientation estimates have low confidence (high variance) at some locations on grain boundaries and in anomalous region at bottom right. This low confidence is quantified by the ML estimator of the scale parameter $\kappa$ of the VMFm model, shown in Fig.~\ref{fig:indexing_uncertainty} for $k=4$ and $k=10$.

\section{Conclusion and Future Work}
\label{sec:conclusion}
We have introduced a novel method for indexing polycrystalline materials that uses both mathematical-physics modeling and mathematical-statistics modeling. The physics-based forward model is discretized into a dense dictionary of diffraction patterns that are indexed by Euler angle triplets. The dictionary is fixed for each crystal symmetry group and each SEM instrument configuration. The statistical model is based on the group symmetry of quaternion representation of the Euler angles on the $3$-sphere in $4$ dimensions. A feature of this method is that it performs classification, segmentation, and indexing in the unified framework of dictionary matching. A feature of the indexing method is that it incorporates a concentration parameter that can be estimated jointly with the Euler angles of a pixel or of a grain. This concentration parameter can be used to report the degree of confidence one can have in the Euler estimates. An iterative maximum likelihood estimator was proposed for estimating the orientation and associated confidence parameters in a statistical Von Mises-Fisher mixture model. The method was illustrated on a single sectional slice of a Nickel alloy sample.

As the proposed indexing method is pixel driven it is directly applicable to indexing over $3$ dimensional volumes. Future work will include algorithm acceleration to make full volumetric indexing fast enough to be practical. Potential acceleration methods include multi-resolution and multi-scale trees, fast coordinate ascent ML optimization, and parallelization. Other areas for future work include robustification of the dictionary to model mismatch, sensitivity to reductions in detector image resolution, and extensions of the dictionary approach to other electron diffraction modalities, such as electron channeling patterns and precession electron diffraction. Preliminary investigations indicate that, while dictionary-based classification appears to be robust to model mismatch, the proposed dictionary-based indexing algorithm is somewhat sensitive to model mismatch. This suggests that the dictionary design may need to be fine tuned to the SEM instrument in addition to the sample’s crystal symmetry group. The extension to other SEM modalities such as EDS is also possible but would require development of dictionaries that capture other types of data (e.g., spectra).

\section*{Acknowledgements}
The authors thank Megna Shah of Bluequartz software for having provided the IN100 sample. We also  thank Michael Groeber  and Michael Uchic at AFRL for their comments on this work.  The CMU portion of this work is supported by the Office of Naval Research on contract no.\ N00014-12-1-0075.
The UM portion of this work was partially supported by USAF/AFMC grant FA8650-9-D-5037/04 and by Air Force Office of Scientific Research grant FA9550-13-1-0043.

\bibliographystyle{unsrtnat}
\bibliography{MandM2015}

\end{document}